\documentclass[10pt,journal,compsoc]{IEEEtran}

\ifCLASSOPTIONcompsoc
\usepackage[nocompress]{cite}
\else
\usepackage{cite}
\fi
\usepackage[nocompress]{cite}
\usepackage{graphicx}
\usepackage{amsmath,amsthm}
\usepackage{verbatim}
\usepackage{caption}

\usepackage{cite}

\usepackage{graphicx}
\usepackage{amsmath,amsthm}
\usepackage{algorithm}
\usepackage{algorithmicx}
\usepackage{algpseudocode}
\usepackage{amsmath}

\algnewcommand\algorithmicforeach{\textbf{for each}}
\algdef{S}[FOR]{ForEach}[1]{\algorithmicforeach\ #1\ \algorithmicdo}

\usepackage{subfigure}
\ifCLASSINFOpdf
\else
\fi
\begin{document}
%
\title{Coalition formation for Multi-agent Pursuit based on Neural Network and AGRMF Model }
%
%
%
%

\author{Zhaoyi Pei,
        Songhao Piao,
        Mohammed Ei Souidi
\IEEEcompsocitemizethanks{\IEEEcompsocthanksitem Songhao Piao was with the
  School of Computer Science, Harbin Institute of Technology, Harin,China,150000.\protect\\
E-mail: piaosh@hit.edu.cn
}
}
\markboth{Journal of \LaTeX\ Class Files,~Vol.~14, No.~8, August~2015}%
{Shell \MakeLowercase{\textit{et al.}}: Bare Advanced Demo of IEEEtran.cls for IEEE Computer Society Journals}
%



\IEEEtitleabstractindextext{%
  \begin{abstract}
    An approach for coalition formation of multi-agent pursuit based on
    neural network and AGRMF model is proposed.This paper
    constructs a novel neural work called AGRMF-ANN which consists of feature extraction part
    and group generation part. On one hand,The convolutional layers of feature
    extraction part can abstract
    the features of agent group role membership function(AGRMF) for all of the groups,on the other
    hand,those features will be fed to the group generation part based on
    self-organizing map(SOM) layer which is used to group the pursuers with similar
    features in the same group. Besides, we also come up the group attractiveness function(GAF)
    to evaluate the quality of groups and the pursuers contribution in order to
    adjust the main ability indicators of AGRMF and other weight of all neural
    network. The simulation experiment showed that this proposal can improve the effectiveness of
    coalition formation for multi-agent pursuit and ability to adopt
    pursuit-evasion problem with the scale of pursuer team growing.

  \end{abstract}

  \begin{IEEEkeywords}
    multi-agent system;neural network;AGRMF;pursuit-evasion
  \end{IEEEkeywords}}

\maketitle

\IEEEdisplaynontitleabstractindextext

%
\IEEEpeerreviewmaketitle

\ifCLASSOPTIONcompsoc
\IEEEraisesectionheading{\section{Introduction}\label{sec:introduction}}
\else
\section{Introduction}
\label{sec:introduction}
\fi
Nowadays, multi-agent systems(MAS) attract more attention in research concerning the
distributed artificial intelligence, Many research on this subject findings have been applied to
intelligent robots, and have achieved amazing results. NASA is exploring a
multi-agent system known as swarmie for the exploration of extraterrestrial
resources\cite{Duncan}, each agent has a complete set of communication devices
and sensors, it can independently discover
potential value resources which can be explored by the partners they call after
that. Such a system has a certain degree of robustness, even if part of it was
damaged it would not be all paralyzed. The adaptive system research team of
Harvard University designed a multi-robot system in 2014 to achieve complex
behaviors such as self assembly just relying on the mutual communication between
adjacent agents\cite{Michael}. It is an ideal platform for studying complex
systems and group behavior.The Autonomous Vehicle Emergency Recovery Tool(AVERT)
studies the tasks which need agents closely coordinated to accomplish such as
lifting weights, pushing boxes, etc. four robots based that tool can easily move and transport two tons of vehicles through collaboration\cite{Richard}.
\par These applications demonstrate the state of art of MAS.Pursuit problem is a
growing field in MAS, with many applications such as search and rescue in
disaster area, patrol on a larger scale and military combat.cooperation
formation of pursuers(agents hunting the evaders) is one of the key technologies of pursuit
evasion. Because when multi-agent pursuit systems are facing multiple targets, it
is bound for them to deal with the problem regarding the optimization of
pursuers tasks coordination. it is optimization problem. For a multi-agent
system, how to make all the pursuers turn material resources to good
account relates to the difference of pursuers as well as that of evaders, and
environmental dynamics, This problem is considered as NP-hard
problem and many novel and effective approach are proposed to solve this
problem, one of which is to learn the each agent's attributes during the process
of pursuit such as emotion model\cite{de}, Threshold method\cite{Gage}. Mohammed
El Habib Souidi, Piao Songhao came up a cooperation mechanism called agent group
role membership function(AGRMF) model in 2015.\cite{Souidi} AGRMF learned from environment and the process of pursuit can reflect
the membership between a pursuer and a group inspired by fuzzy logic
principles, so that MAS will assign the tasks according to AGRMF instead of
assigning them to any agent without considering their ability and position. this
approach has proved to be effective, but it will be hard to set the main ability
indicators\cite{Souidi} of each agent when the scale of MAS is too large which means there are
many agents in a relative larger environment, what we want to do is make up for this deficiency.
\par This paper proposes a coalition formation mechanism of multi-agent pursuit
based on neural network and AGRMF model, the pursuers use the special main
ability indicators from AGRMF-ANN in face of given evaders, we apply the novel neural
network AGRMF-ANN to abstract each pursuer's AGRMF feature and learn the main ability
indicators, then the self-organizing map(SOM) fed by those features will
create groups for pursuers,then pursuers in one group will be assigned their target according to AGRMF, in order to adjust the main ability indicators of AGRMF and other weights of all neural
network, we come up the group attractiveness function(GAF) to evaluate the
quality of groups and the pursuers contribution. Finally, we apply the markov decision
process to promote pursuers get their own optimal motion planning after the
formation generated during each iteration.\cite{Souidi}
\par The paper is organized as follows:in the section 2, we will introduce some related work about machine
learning principles applications for multi-agent pursuit, in the section 3, we will describe
the multi-agent pursuit problem in grid environment based on AGRMF model, the part of features
extraction for pursuers and that of group generation will be discussed
respectively in section 4 and section 5. The group attractiveness function will be
deduced in section 6 as well as the network training process. The total
algorithm inspired by our approach is showed in section 7.
\section{Related Work}
There exist a lot of approaches based machine learning processing the
pursuit-evasion problem. Taking into account the Reinforcement learning is
considered as a famous process for solving motion optimization problem in the
environment where agents can get delayed reward. In
order to solve the problem in which the pursuers are at disadvantage regarding
the motion speed in in comparison with the existing evaders, Mostafa D. Awheda and Howard M. Schwartz  use fuzzy reinforcement learning algorithm to learn different multi-pursuer single-superior-evader pursuit-evasion differential games with the capture region of the learning pursuer defined by Apollonius circle mechanism.\cite{Awheda}
Ahmet Tunc Bilgin and Esra Kadioglu-Urtis proper a new approach based on Q
learning with multiple agents characterized by independent learning. In their study, they also give the evaders the ability to learning optimized strategy from the environment.\cite{Bilgin}
Andrew Garland and Richard Alterman apply learning technology which can promote agents to use past runtime experience to solve coordination problems with personal goals, so that their future behavior can more accurately reflect past successes stored in their individual case.\cite{Garland}
To solve the problem regarding the identification of the best strategy to enclose an intruder for pursuers and adopt the huge environment,SHIMADA Yasuyuki and OHTSUKA Hirofumi discussed how to discretize patrol areas based on reinforcement learning.\cite{Yasuyuki}
Francisco Martinez-Gil and Miguel Lozano compared the two different learning algorithms based on Reinforcement learning :Iterative Vector Quantization with Q-Learning (ITVQQL) TS, Tile coding as the generalization method with the Sarsa(TS). The simulation results for  pedestrian showed that the two RL framework had common advantage that they could generate emergent collective behavior which can help them arrive their destination collaborative.\cite{Martinez-Gil}
Khaled M. Khalil and M. Abdel-Aziz designed a software framework for machine
learning in interactive multi-agent systems based on Q-learning approach, this framework provides an interface for users to operate with the
processes of agents learning activities and sharing information.\cite{Khalil}
\par Besides, many other approach such as neural network are also suitable for
solving pursuit-evasion problem based on multi-agent system. Ding Wanga and Hongwen Ma focused on dealing with the
existence of negative communication weights which can effect bipartite consensus
in multi-agent system seriously and propose a distributed control algorithm
based on RBFNN.\cite{Wang}
Jong Yih Kuo, and Hsiang-Fu Yu designed an adopt agent learning model including a special module called belief module which can help a pursuer learn plans from either success or failure experiences in the absence of environmental information.\cite{Kuo}
Lin Zhao and Yingmin Jia applied the Lyapunov approach and graph theory on
fault-tolerant parts included in controllers to compensate the partial loss of
actuator effectiveness faults.\cite{ZYJ Lin}
\section{Description of Multi-agent Pursuit Problem}
\subsection{Pursuers}
The set of pursuers is denoted by $$P = \{p_1,p_2,...p_n\}$$. Each agent should be
described by two additional Capacity parameters in order to determine membership
function of AGR model:
\par
\textbf{Self-conference Degree.} This parameter can reflect the ability of
agents to play the role of pursuers which can be calculated by follows:
\begin{equation}
  \forall Conf \in [0.1,1]: max(0.1,\frac{c_t}{c_s})
\end{equation}
where $C_s$ is the number of tasks that the agent has finished with success
\par
\textbf{Credit.}If there are tasks which the agent failed to finish,its credit
will be influenced,the credit of an agent is denoted as follows:
\begin{equation}
  \forall Credit \in [0.1,1]:min(1,1-\frac{c_b}{c_t-c_s})
\end{equation}
while $C_b$ is the nubmer of tasks that the agent has to abandon.  $C_t$ is the number
of tasks that the agent has participated.
\par distance from pursuer to evader  in environment is a crucial
criterion for the pursuit-evasion,because the pursuit will be easier if the
pursuer is closer to the evader. The distance Dist between pursuer P and evader E is computed as follows:
\begin{equation}
  Dist_{PE}=\sqrt{(cc_{P_i}-cc_{E_i})^2+(cc_{P_j}-cc_{E_j})^2}
\end{equation}
$(cc_{P_i},cc_{P_j})$ is the Cartesian coordinates of the pursuer.
$(cc_{E_i},cc_{E_j})$ is the Cartesian coordinates of the evader.
\subsection{Evaders}
The set of evaders is denoted by $$E=\{e_1,e_2,...e_n\}$$
each evader has its own pursuit difficulty which means how many pursuers are
needed to catch that evader, the set of pursuit difficulty is denoted by $$D=\{d_1,d_2,...d_n\}$$
pursuer who catch evader e will get the reward which equal to $d_e$.
\subsection{Organizer}
\par This agent is the mainstay of the chase as it will create the chase groups
and extract features of pursuers,therefore it is the place where AGRMF-ANN is deployed.The organiser should be trusted by all pursuers and can't
disclose the parameters of pursuer to others.
\subsection{Rule of pursuit}
One evader will be considered as arrested if the adjacent cells are occupied by other agents or obstacles
 as equation \ref{sucarrest} and fig. 1.
\begin{equation}
  d_n \leq \sum_{x \in neighbor(d_n)}{P(x)+O(x)}
  \label{sucarrest}
\end{equation}

Where $neighbor(d_n)$ means the grid cells left to $d_n$ or right to that,on top
of $d_n$ or under that.P(x) means the grid cell $x$ is whether occupied by
pursuers or not and O(x) means whether that is obstacle or not.
\begin{figure}
  \centering
  \includegraphics[width=0.5\linewidth]{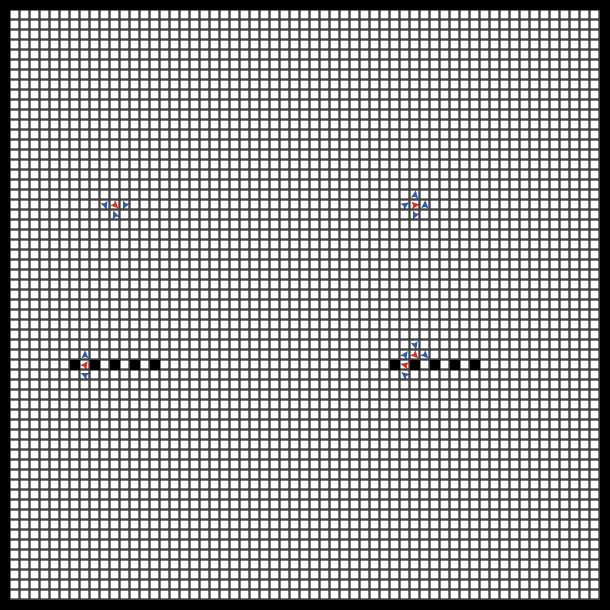}
  \label{successarrest}
  \caption{instances of successful arrest}
\end{figure}
\section{Feature Extraction of Pursuit Agents}
\subsection{agent group role membership function}
Agent group role membership Function(AGRMF) is an extension of agent-group-role
model. Unlike other institutions,this approach allows agents to consider group as
fuzzy set. Consequently, each group can be managed via the utilization of a membership function.
\par  Instead of figuring out one explicit result that whether one special agent belongs
to one of the groups or not comparing with ordinary AGR model, AGRMF admits degrees
of membership to a given group.So that this change will provide
more space to use different algorithm to optimize the result for the organization and more flexibility for
reorganization comparing ordinary AGR model. Roles of one group can not be played
by any agent without considering its property. The AGRMF($\mu_{Group}(Agent)$)
represents mission completion success probability as equation \ref{membershipfunction}.
The basic sequences of the coalition generation algorithm based on AGRMF model which is proposed by
literature\cite{Souidi2} is shown as algorithm 1.
\begin{equation}
  \mu_{Group}(Agent) =\frac{Coef_1*Pos+Coef_2*Conf+Coef_3*Credit}{\sum_{i=1}^3{Coef_i}}
  \label{membershipfunction}
\end{equation}
\begin{algorithm}
  \small
  \scriptsize
  \caption{getgroupofpursuers}
  \begin{algorithmic}[1]
    \Require $U$ the set of vector of AGRMF of evader ,$n$ size of $U$
    \Ensure the list consisting of the set of pursuers take responsible for each pursuers
    \Function {getgroupofpursuers}{$U,n$}
    \State pursuersgorup:empty dictionary
    \ForEach {$e \in E$}
    \State x$ \gets 0$
    \Repeat
    \ForEach {$p \in P$}
    \If {$(p \not \in pursuergroup[e]) \&\& u_e(p) = Max(u_e) $}
    \State Add (pursuersgorup[e],p);
    \State Delete (P,p);
    \State $x \gets x + 1$
    \EndIf
    \EndFor
    \Until {x=e.d}
    \EndFor
    \EndFunction
  \end{algorithmic}
\end{algorithm}
\par
The main idea of this algorithm is to select the pursuer with the largest
membership degree of target to join group. AGRMF is the
most important part of the whole algorithm. the main ability
indicators($Coef_1-Coef_3$) which can reflect the contribution of each factor to
the $\mu_e$ and probability of pursuit success is critical parameters of each
evader,the set of each evader own main ability of indicators will be hard to define
when there are a large number of agents in large environment and it is not
appropriate to set them as constants because of the variability of the
environment.We will discuss the feature extraction part of our neural network
which trys to solve this problem in the remainder of this section.
\subsection{Convolutional Neural Network}
BP network is a kind of multi-layer network which uses back propagation algorithm to learn from train set,the error of one layout will be propagated back to the
previous one ,so it will be calculated until it is propagated back from
output layout to input layout,the weights' change is relay on the error
of the layout which they belong to in order to make sure the output of the whole network approach the
expectation.A fascinating characteristic of BP network is that it is able to find useful intermediate representations within the hidden layer inside the network.
Any function can be approximated by a neural network with three layers with any
small errors.\cite{Cybenko}. But as far as practical applications are
concerned,When the amount of data is large and the features of the samples are
complex,It's a good idea to make the neural network thinner(that means more
layers and less neurons of each layers),So convolutional Neural Network appeared.
\par
\begin{figure}
  \begin{minipage}[t]{0.5\linewidth}
    \centering
    \includegraphics[width=1.5in]{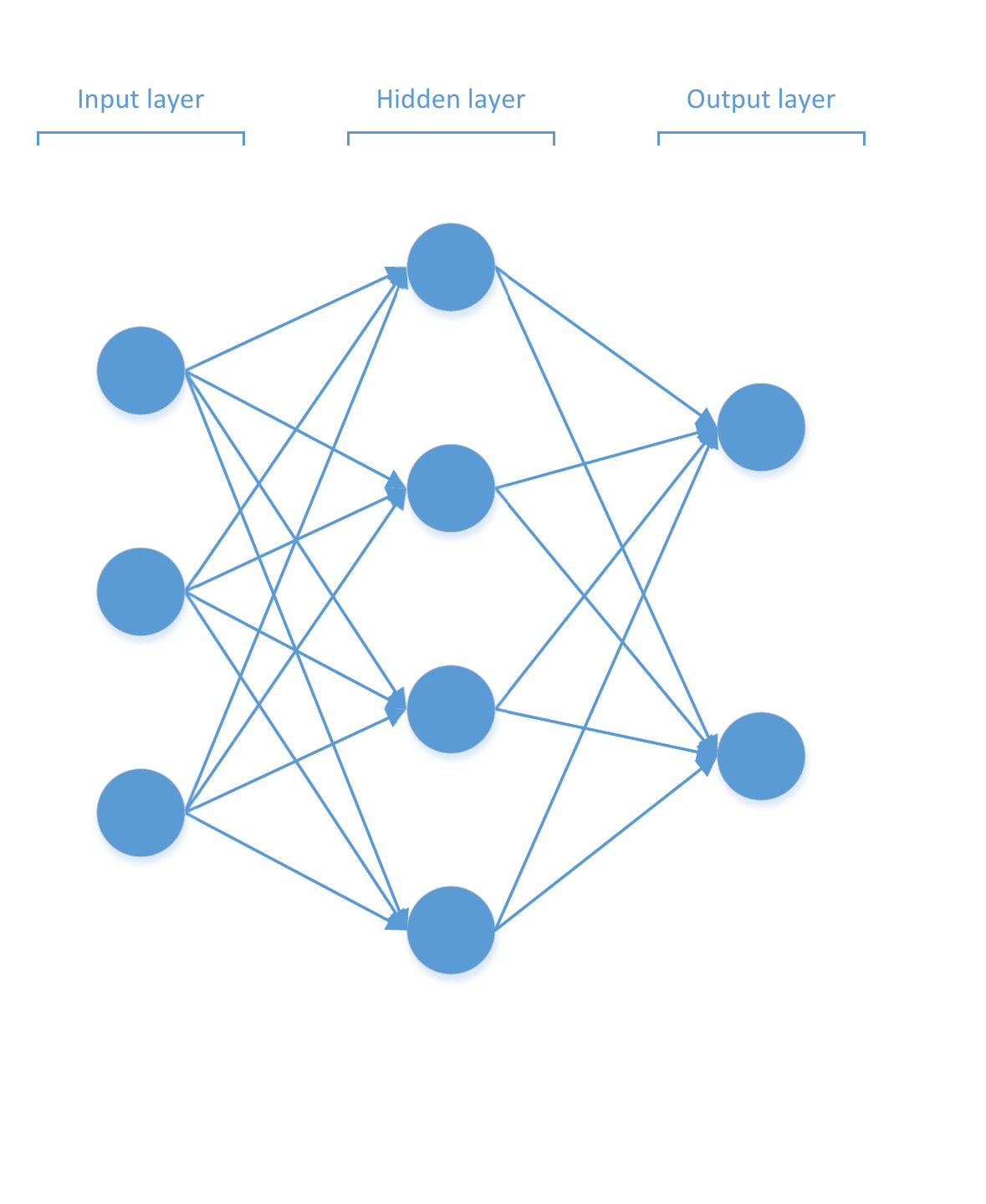}
    \caption{\label{bpnetwork} BP neuron network}
  \end{minipage}%
  \begin{minipage}[t]{0.5\linewidth}
    \centering
    \includegraphics[width=2.0in]{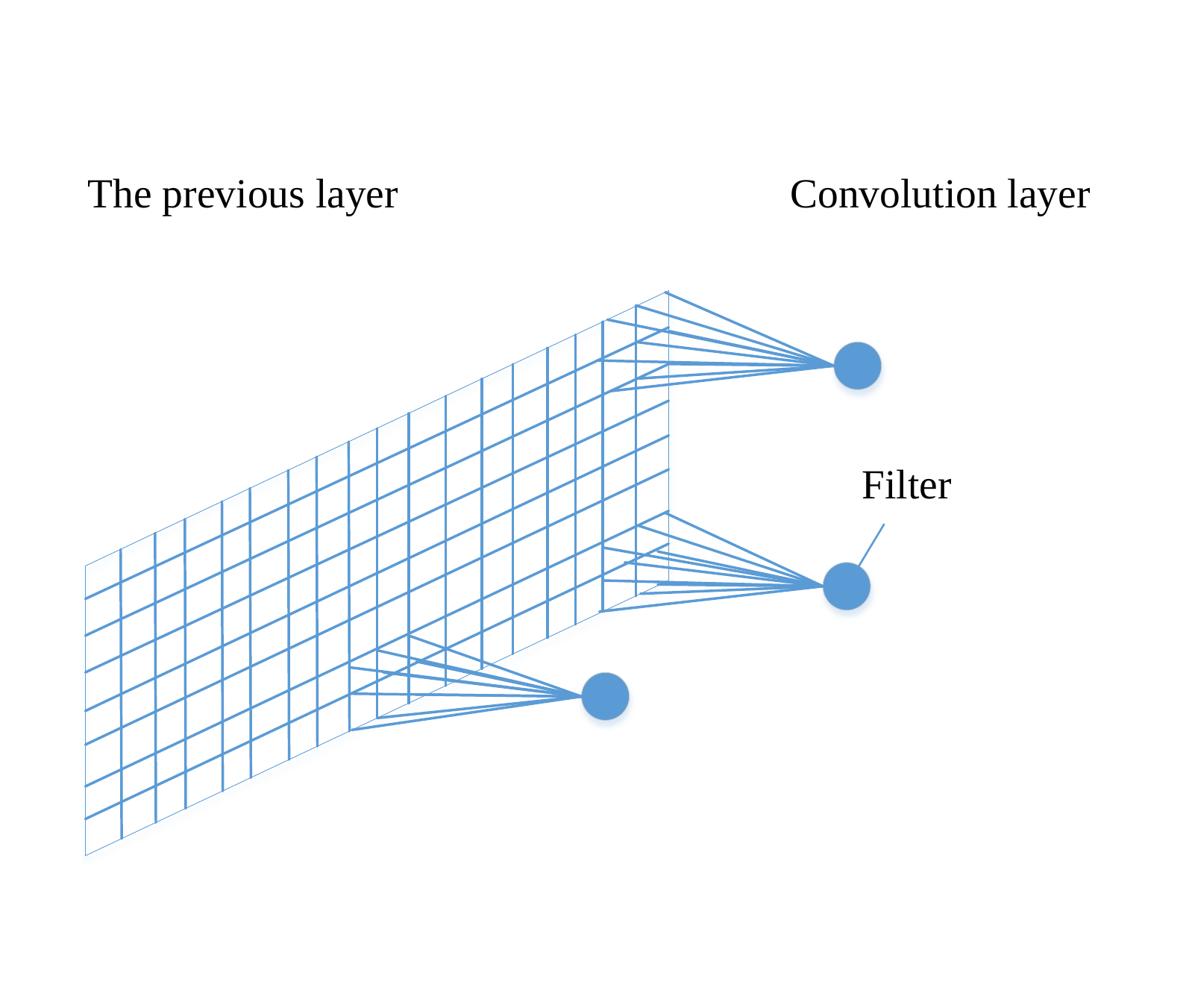}
    \caption{convolution neuron network }
    \label{CNN}
  \end{minipage}
\end{figure}

 CNN was proposed by leCun in literature\cite{leCun} in 1989. Nowadays, CNN
is widely applied image recognition, speech processing, artificial intelligence.
The biggest difference between standard BP and CNN is that in CNN neurons
adjacent layers are partial connected instead of fully connected,In other words,a
neuron's sensing area comes from part of the previous layer as figure \ref{CNN}
and figure \ref{bpnetwork}, rather than connect
all neurons.The neurons of convolutional layers are also called filters.The formula for the forward calculation of the convolution layer are as follows:
\begin{equation}
  x^l_j=f(\Sigma_{i \in M_j}X^{l-1}_i*k^l_{ij}+b^l_j)
\end{equation}

\par The essence of the above is to let the convolution kernel k do convolution
on all associated feature maps in layer $L-1$, and then sum up, plus a
bias parameter,obtain the final excitation value inspired by  activation function.
\subsection{Feature Extraction of Pursuers}
\begin{figure}
  \centering
  \includegraphics[width=0.8\linewidth]{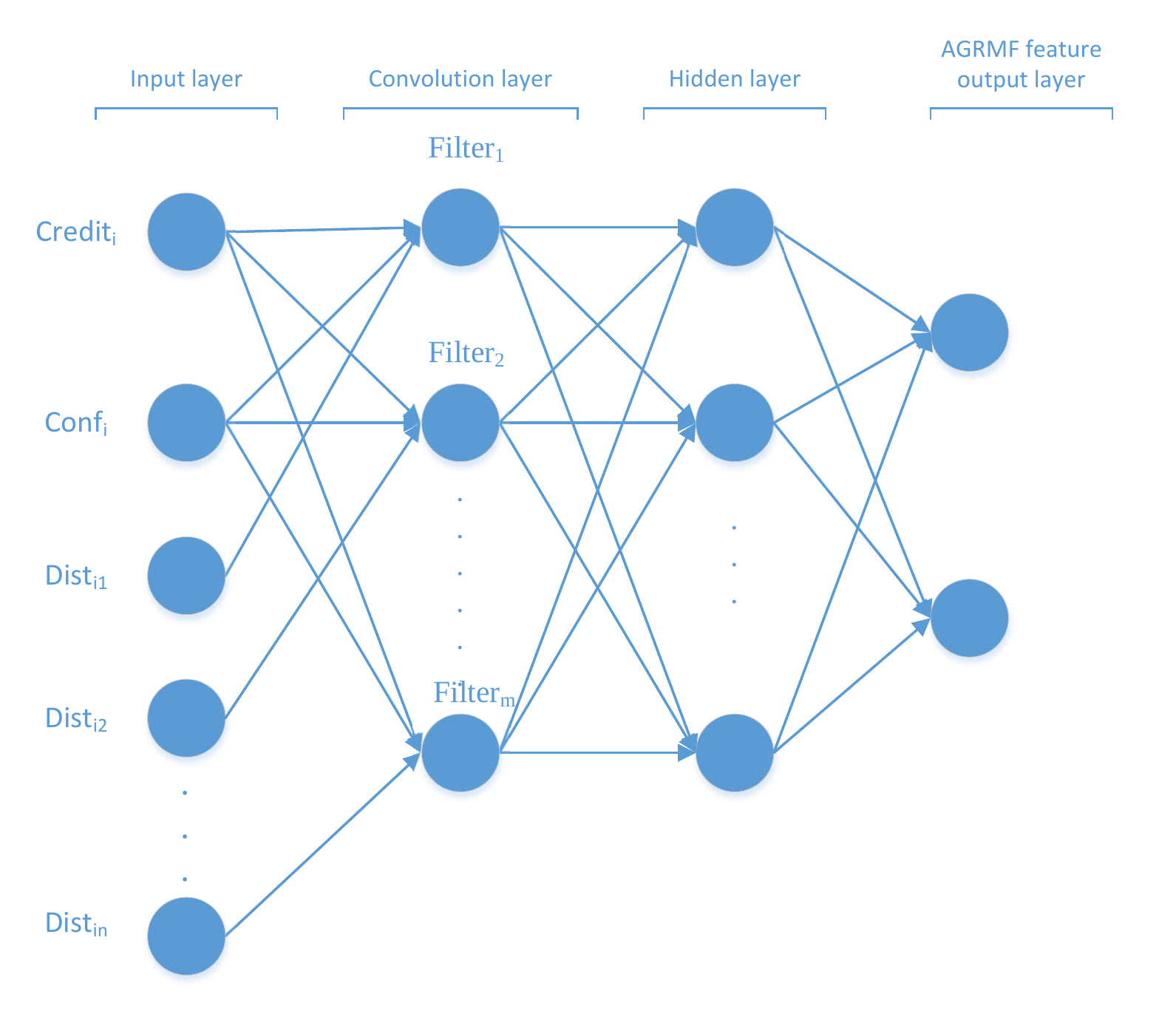}
  \caption{feature extraction part for pursuers}
  \label{ournetwork}
\end{figure}

In this section,we will discuss the feature extraction part for pursuers of our neuron network,
pursuer $i$ can be represented by  one feature vector
$\overrightarrow{x_i}=\{Cedit_i,Conf_i,Dist_{i1},Dist_{i2}
...,Dist_{im}\}$,where the $Dist_{im}$ means the distance between pursuer $i$ and evader $m$.The input layer is followed by the
convolutional layer. Each neuron of the convolutional layer is the filter
of each pursuer, the filter of evader i just connects with
$Credit_i$,$Conf_i$,$Dist_{ij}$ By means of convolution operations, stimulus of
$filter_j$ which also represents the $\mu_j(i)$ is obtained by $x_i$, The weight
of the filer represents the contribution rate of each attribute to the
membership of the group and they can be regarded as main ability indicators, so
the $Coef_1,Coef_2,Coef_3$ in equation \ref{membershipfunction} will be replaced
by $w_1,w_2,w_3$ of filters. Then, the hidden layer can improve the approximation ability of the neural network.The structure of this part is shown in Figure \ref{ournetwork}.
\subsection{Manual Data Synthesis}
The number of input vectors collected by the method for each iteration proposed in section 4.3 is
$|P|$ which is not sufficient for training the neural network,We notice
that the neighborhood of the pursuer can also well represent the characteristics
of the agent as fig \ref{runtest3} when the scale of environment is large,the number of data can be increased from $|P|$ to $|P|*|Neighborhood|^{|E|+1}$
\begin{figure}
  \centering
  \includegraphics[width=0.5\linewidth]{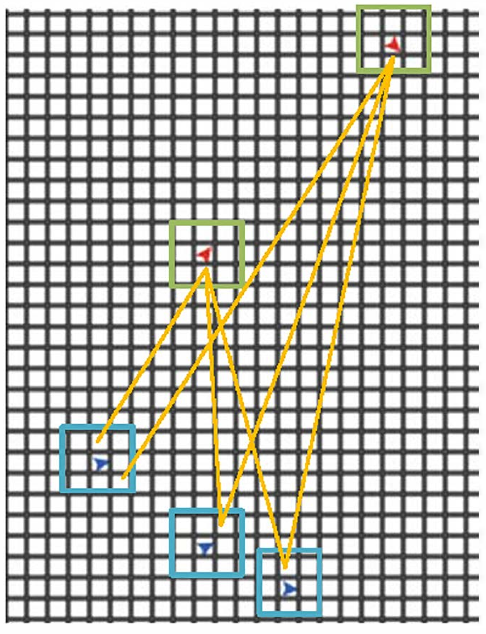}
  \caption{Manual data synthesis}
  \label{newdata}
\end{figure}
\section{Groups Generation with Feature of Pursuit Based on SOM}
\subsection{Self Organized Map}

The Self Organized Map is one of the most popular neural network
models which is a type of unsupervised learning approach\cite{IM} ,
SOM is widely applied for clustering data without knowing the class
category of the input data.\cite{Vesanto}
the SOM is based on competitive learning network.Competitive learning network
excavates characteristics of the input vector and cluster them.Provided training
set as $\{X^1,X^2,X^3...X^p\}$ without pre-assigned label, for each $X$ in
training set, the network outputs one winner node after competitive learning
,not like the BP network outputs amplitude, the vector connected to
the winner node will represent the X's information best in the whole
network, when one $X$ input the network, there must be one node get the peak
response, so the competitive learning is the process to make the winner node to
get the peak response, output node $i$ connects input node $j$ with $w_{ij}$, for
each $X^p$, define the response of output node i as equation \ref{somequa1}
\begin{equation}
  \sum_{j=1}^n W_{ij}*X_j= \stackrel{\longrightarrow}{W_j}^T*\stackrel{\longrightarrow}{X}
  \label{somequa1}
\end{equation}
$y_i$ will be the winner node if it as equation \ref{somequa2}.
\begin{equation}
  y_i=max_{k=1,2,3...m}(\stackrel{\longrightarrow}{W_k}^T*\stackrel{\longrightarrow}{X})
  \label{somequa2}
\end{equation}
which mean the node i get the peak response at this time, after normalizing the weights it is equivalent to expression \ref{somequa3}.

\begin{equation}
  \left | W_i-X_i \right|=min(\left | W_i-X_i \right|)
  \label{somequa3}
\end{equation}
\begin{figure}
  \begin{minipage}[t]{0.5\linewidth}
    \centering
    \includegraphics[width=2in]{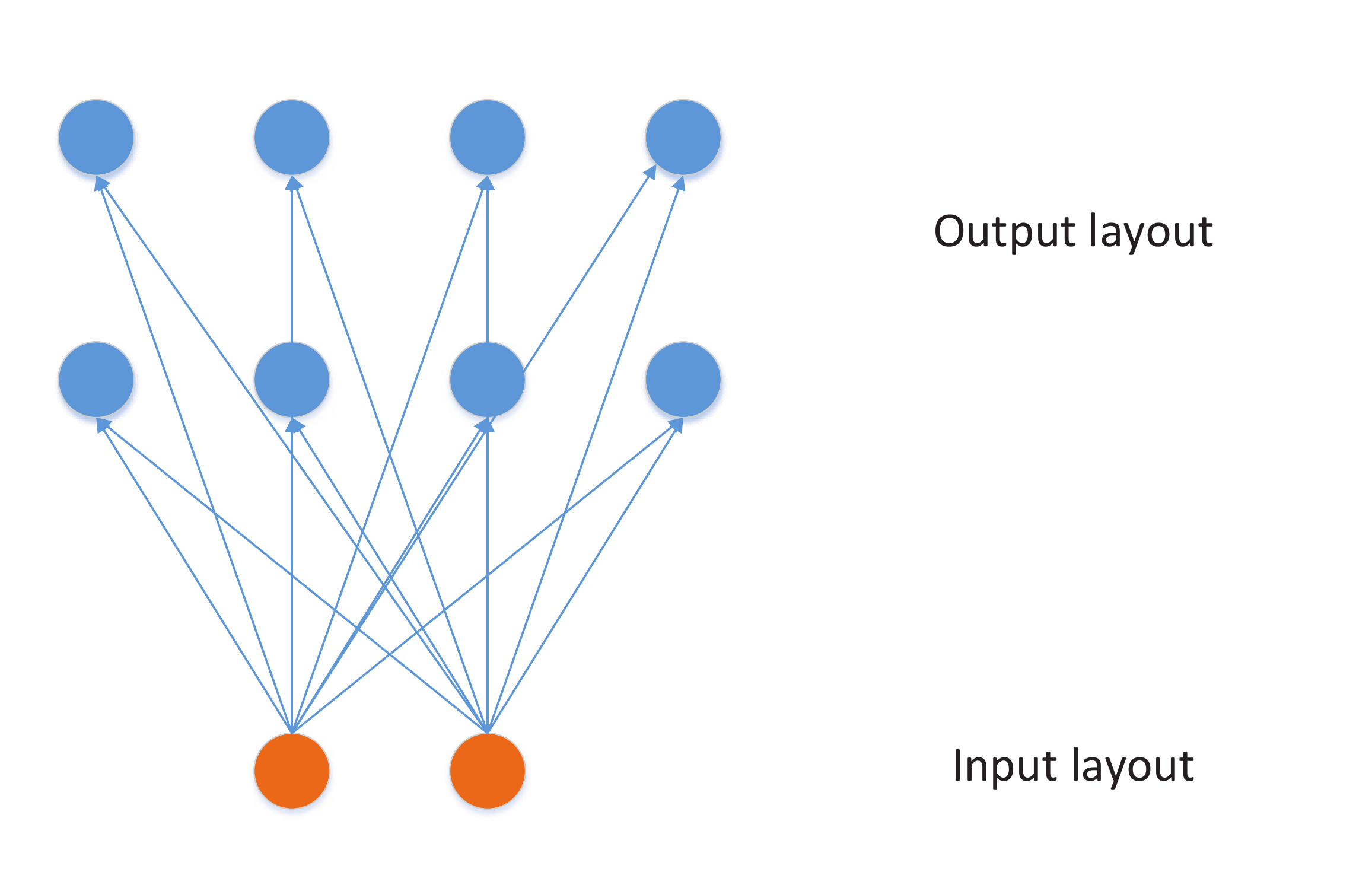}
    \caption{\label{structure of SOFM} structure of SOFM}
  \end{minipage}%
  \begin{minipage}[t]{0.5\linewidth}
    \centering
    \includegraphics[width=1.5in]{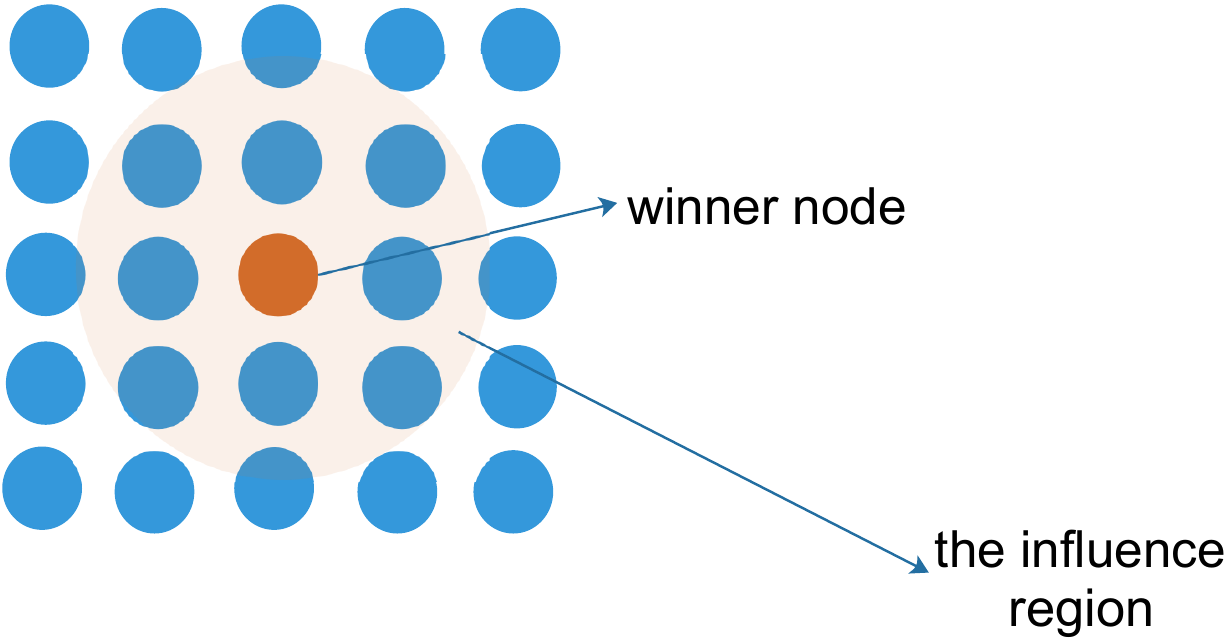}
    \caption{influence region of winner neuron }
    \label{influence region of winner neuron}
  \end{minipage}
\end{figure}

The ordinary competitive learning network is enough to deal with the
representative cluster problem,however, it is hard for it to deal with the huge
input data set, the SOFM can overcome this shortcoming, \cite{Nakao} besides, making use of the
topological structure of SOM, the relation between data can be clearly
visualized.\cite{Mohebi} A SOM maps the neurons upon one matrix of 2-dimensional or
3-dimensional generally just like fig \ref{structure of SOFM}. SOFM is based on
the competitive learning network but each neuron of output layout is connected
closely with the neighboring neurons not like the basic competitive learning
network where each neuron is independent. The connections is called lateral
connections, each neuron has its own region of influence for each iteration as
fig. \ref{influence region of winner neuron}.
Each neuron compete with the neighboring neurons via lateral connections, the
winner neurons will get the strongest inspirit and the neighboring neurons in
its influence region will get the different inspirit or inhibition. The relative
close neurons in the influence region will be inspirited by the winner neurons
but the others be suppressed, the neurons out of the region will have no influence.
By this way, the competition will be more effective, and the equation of weights' update will be changed as equation \ref{updateweight}.
\begin{equation}
  \left \{ \begin{matrix}
      w_{ji}(t+1) =w_{ji}(t)+\alpha (t)(x_i(t)-w_{ji}(t)) & j \in N_c(t) \\
      w_{ji}(t+1)=w_{ji}(t)  & j \not \in N_c(t)
    \end{matrix} \right.
  \label{updateweight}
\end{equation}

\begin{figure}
  \begin{minipage}[t]{0.5\linewidth}
    \centering
    \includegraphics[width=1.5in]{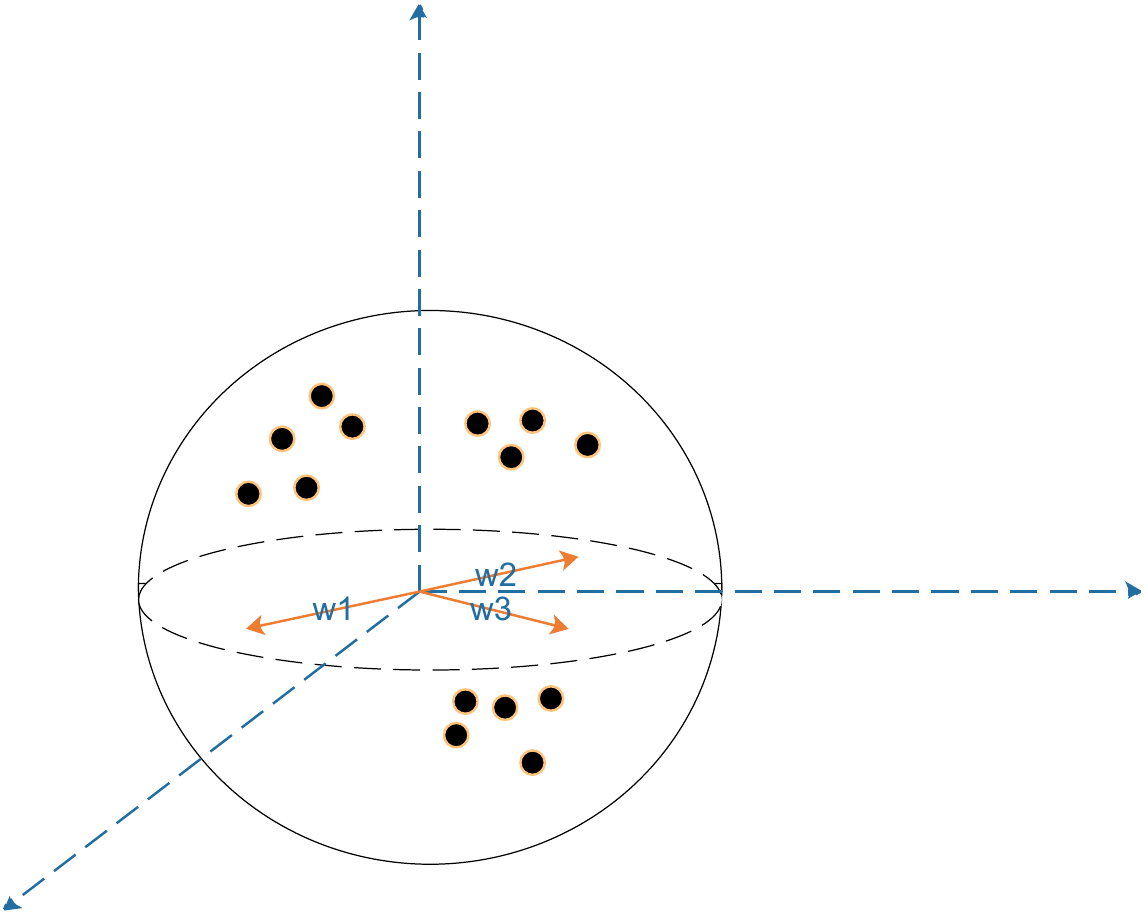}
    \caption{\label{network before trained} network before trained}
  \end{minipage}%
  \begin{minipage}[t]{0.5\linewidth}
    \centering
    \includegraphics[width=1.5in]{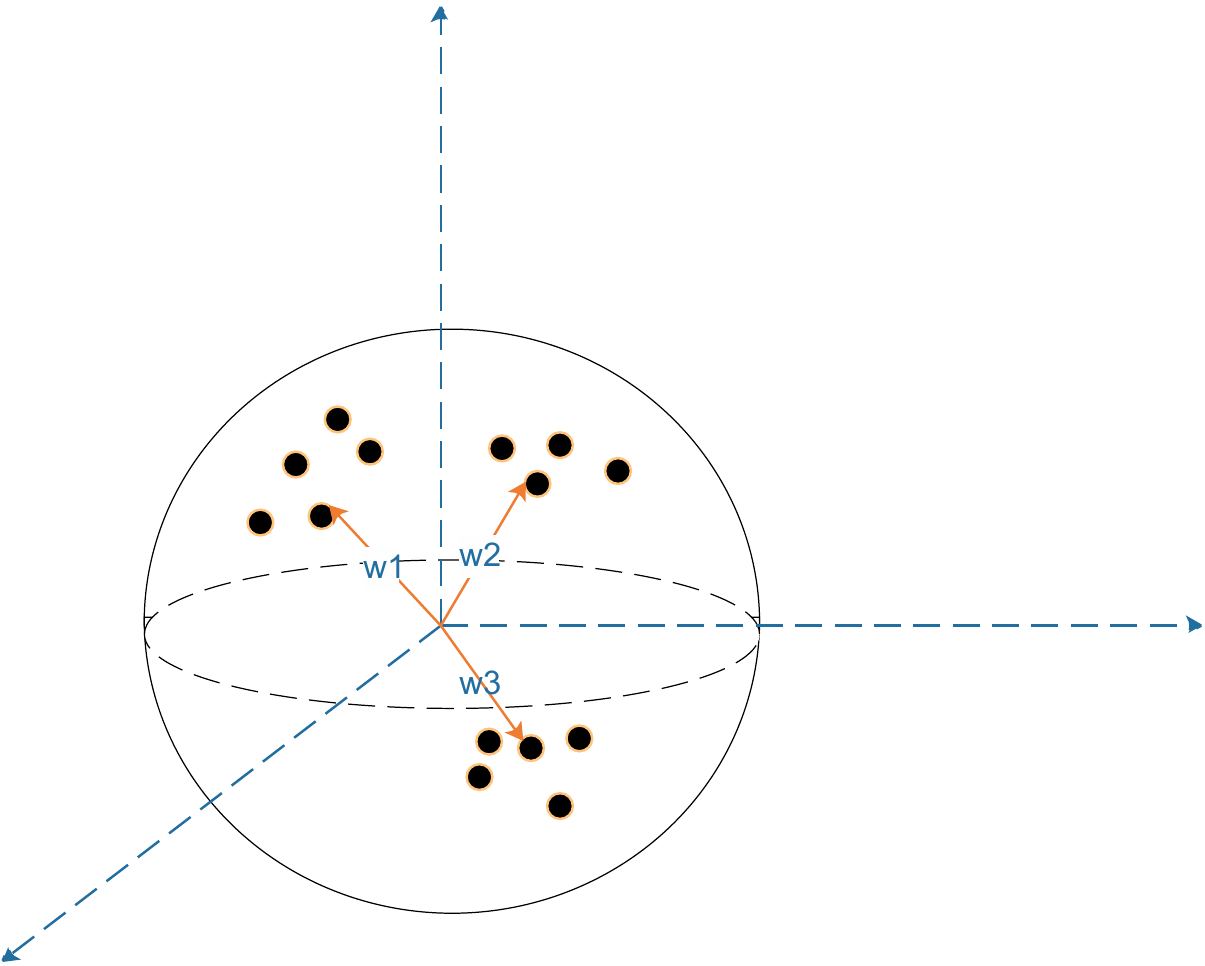}
    \caption{network after trained }
    \label{network after trained}
  \end{minipage}
\end{figure}

The $\alpha(t) $ is learning rate, and $N_c(t)$ is the region of influence, where
$t$ is the the number of current iteration they will change as equation
\ref{alpha} and equation \ref{Nt} to make sure the network will achieve convergence.
\begin{equation}
  \alpha (t)=\alpha (0)[1-\frac{t}{T}]
  \label{alpha}
\end{equation}
\begin{equation}
  N_c(t)=INT[N_c(0)(1-\frac{t}{T})]
  \label{Nt}
\end{equation}
Actually,the competitive learning is the process of the weights of
node to approach the center of group, just as fig.\ref{network before
  trained} and fig.\ref{network after trained} where the input space is a
2-dimensional space, $w_1$ ,$w_2$ and $w_3$ are weight vectors .
\subsection{SOM Layer for Group Generation}
\begin{figure}
  \centering
  \includegraphics[width=0.8\linewidth]{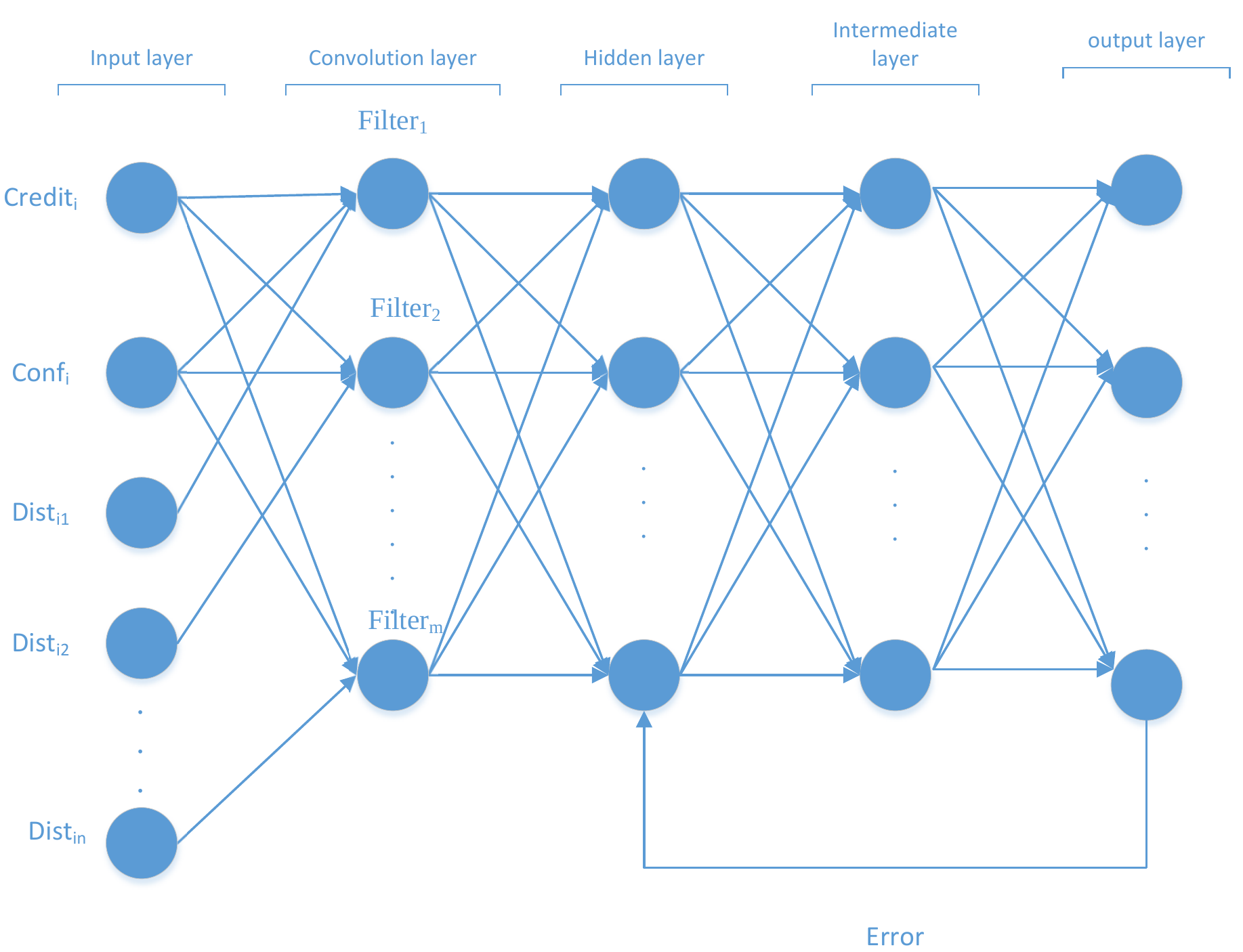}
  \caption{neuron network for Coalition Formation}
  \label{ournewnetwork}
\end{figure}

The idea of our Generation group algorithm is to assign the pursuers with
similar AGRMF feature vectors to the same group without requiring predefined number
of group. Therefore, the SOM layer is added after feature
extraction part. The complete construction of neuron network for coalition
formation is shown as figure \ref{ournewnetwork}. The intermediate layer is
output layer of AGRMF features as well as input layer of Generation Groups,The
weights connecting the intermediate layer and output layer updates as equation
\ref{updateweight}. Error is passed from the output layer to the hidden layer
which will be discussed in section 6. The weight vector of each neuron can fully represent the
center of AGRMF features of pursuers whose winner neuron is that neuron When the
neural network converges.
The algorithm’s steps of training the SOM inspired by AGRMF feature vectors and getting
final group result is presented as algorithm 2.
\begin{algorithm}
  \small
  \scriptsize
  \caption{getgroupofevaders}
  \begin{algorithmic}[1] 
    \Require $FU$ the set of AGRMF feature vectors of pursuers, $n$ size of $FU$
    \Ensure the list consisting of index of group each pursuer's belongs to
    \Function {getgroupofpursuers}{$FU,n$}

    \State List:empty list
    \State $\lambda$ :time of iteration
    \State $m$:the number of output neuron
    \For {j 0 to m}
    \State random initialize $W_j$
    \EndFor
    \Repeat
    \State $k \gets 0$
    \Repeat
    \State $i \gets 0$
    \State $f\mu_i \in FU $
    \State $o:=win(W,f\mu_i)$
    \State $updateweight(o)$
    \State $i \gets i + 1$
    \Until i = n
    \State $k \gets k + 1$
    \Until k = $\lambda$
    \Repeat
    \State $i \gets 0$
    \State $f\mu_i \in U $
    \State $o:=win(W,f\mu_i)$
    \State add(o,List)
    \State $i \gets i + 1$
    \Until i = n
    \State return List
    \EndFunction
  \end{algorithmic}
\end{algorithm}
\par The algorithm’s steps are explained in the following manner:the step 2 to 7 is
used to initialize the parameters and SOM, the step 8 to 18 is the learning
process of SOFM where step 13 is inspired by equation 9 and step 14 is inspired
by equation 10. Finally the list consisting of index of group each pursuer
belongs to will be generated. The pursuers who get the same output from SOFM will be assigned to
the same group,the pursuers of group $g$ decides their target evader $e$
according to their AGRMF as equation \ref{getevader}.each group will only own
one target at the same time.Groups that have the same target will be merged into one group.
\begin{equation}
  e = max_{arg}\Sigma_{p \in g }{(\mu_{e_1}{(p)}, \mu_{e_2}{(p)} ... \mu_{e_n}{(p)})}
  \label{getevader}
\end{equation}
\section{Training algorithm of our neural network}
\subsection{group attractiveness function}
Group attractiveness function(GAF) is proposed to evaluate the correctness of
the coalition formation obtained by the neural network.Assigning a pursuer to a
group that is attractive to it will be considered as a correct decision. Group
$g$'s GAF for $p'$ will be directly proportional to the difficulty of its target $d$, the AGRMF $\mu_{p'}(G)$, it is also
inversely proportional to the sum of the distances from the pursuers to the
target which will represents the difficulty of pursuit at current status.
the $GAF_g(p')$ is shown as equation. \ref{GAF}
\begin{equation}
  GAF_g(p')=\frac{\mu_{g}(p')*d}{\Sigma_{p \in g}dis(p,E)}
  \label{GAF}
\end{equation}
It is beneficial to the promotion of the efficiency of the MAS that
each pursuer leaves the original group to pursue the growth of its own
interests\cite{Souidi3},that is the meaning of reorganization.Each agent has to
make a decision between the interests in the original group $GAF_g(p)$ and the
interests in the other groups  $\frac{\Sigma_{g \in
    \{G-g\}}(GAF_{g}(p))}{|G-g|}$,the result also are determined by inertial
factor $c_p$ which represent the nature of pursuers to keep in origin group,This
factor between 0 and 1 can be used to control the speed of the reorganization.The coalition
evaluation function(CEF) is shown as equation\ref{CEF}.
\begin{equation}
  \left \{ \begin{matrix}
      CEF(p) = 1 & C_p*GAF_g(p)>(1-C_p)\frac{\Sigma_{g \in \{G-g\}}(GAF_{g}(p))}{|G-g|} \\
      CEF(p) = 0 & C_p*GAF_g(p)<(1-C_p)\frac{\Sigma_{g \in \{G-g\}}(GAF_{g}(p))}{|G-g|}
    \end{matrix} \right.
  \label{CEF}
\end{equation}
\begin{figure}
  \begin{minipage}[t]{0.5\linewidth}
    \centering
    \includegraphics[width=1.5in]{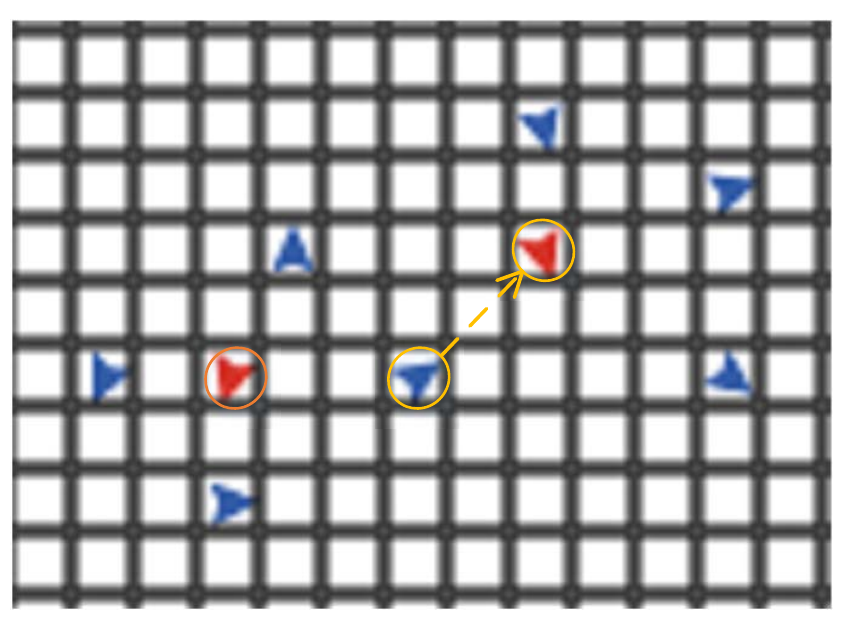}
    \caption{\label{beforeorganization} status before reorganization}
  \end{minipage}%
  \begin{minipage}[t]{0.5\linewidth}
    \centering
    \includegraphics[width=1.5in]{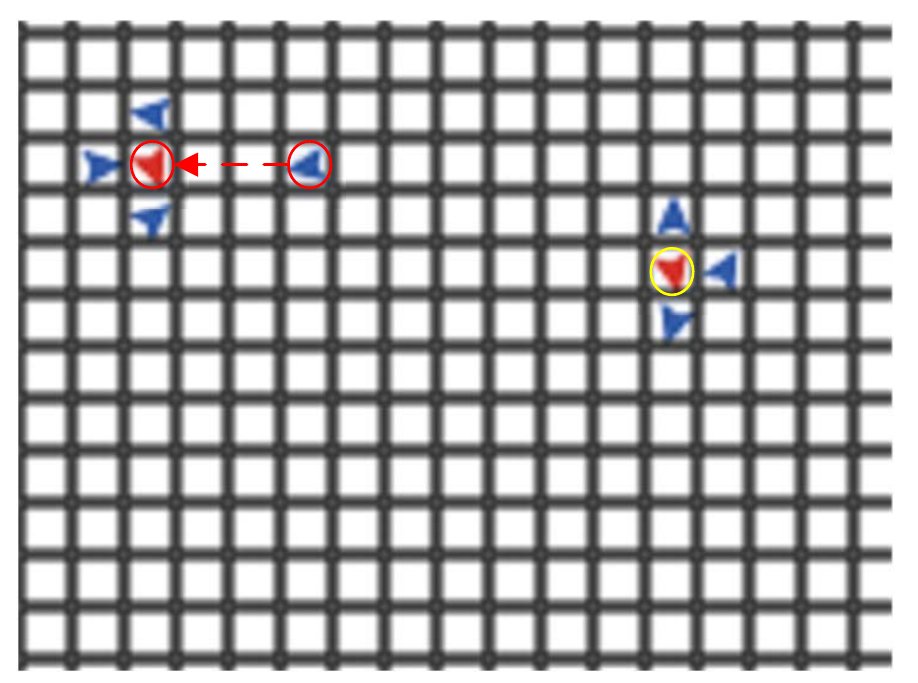}
    \caption{status after reorganization}
    \label{afterorganization}
  \end{minipage}
\end{figure}
The role of CEF is shown in figure \ref{beforeorganization} and \ref{afterorganization}.there are two target with same
value, The left target need 4 pursuers to capture but only 3 pursuers was suit
to be assigned to catch it,lack of pursuer will cause the loss of this target,on the contrary,3 pursuers are enough
to catch the right target though there are 4 pursuers including the one with yellow circles.
Suppose the inertial factor of that pursuer is 0.5.CEF will inply that agent
whose AGRMF  reached a certain standard to leave origin group and join another group this will be considered a better
solution.coalition formation will be translated into classification problem
with the effect of CEF,how to train AGRMF-ANN  will be discussed in the
next subsection.
\subsection{Back propagation based on CEF}
The learning algorithm of AGRMF feature extraction is a new back-propagation algorithm
based on CEF,the label for each train sample of pursuer p can be defined according to CEF'(p)
which is calculated in last iteration.So the loss function can be shown as
equation \ref{lossfunction}.
\begin{multline}
  E(\overrightarrow{w})=\frac{1}{2}(CEF(p)\sum_{d \in D}\sum_{k \in output}^q(win_k(d)-o_k(d))^2+\\ (1-CEF(p))(\sum_{d \in D}\sum_{k \in output}^q(win'_k(d)-o_k(d)))^2
  \label{lossfunction}
\end{multline}
d is the feature vector of pursuer p, $o_k(d)$ is the output vector of AGRMF
feature extraction part as well as the input vector of SOM layer for $o_k(d)$. $w_o(k)$ is the weights of the winner
neuron in SOM layer, $w'_o(k)$ is the weights of the neuron of group g' which
is $max_{arg}GAF(p)$ ,CEF'(p) can be treated as a constant because it is a part
of label and can not be changed at current iteration,$E(\overrightarrow{w})$ is
a continuous derivatived function so that the back propagation algorithm can be
used to update the weights of feature extraction part of AGRMF-ANN as equation \ref{updatebp}.

\begin{multline}
  \delta(w)=-\eta \nabla_w\frac{1}{2}(CEF(p)\sum_{d \in D}\sum_{k \in output}^q(win_k(d)-o_k(d))^2+\\ (1-CEF(p))(\sum_{d \in D}\sum_{k \in output}^q(win'_k(d)-o_k(d)))^2
  \label{updatebp}
\end{multline}
\section{Sequences of our coalition algorithm}
\begin{figure}
  \centering
  \includegraphics[width=0.6\linewidth]{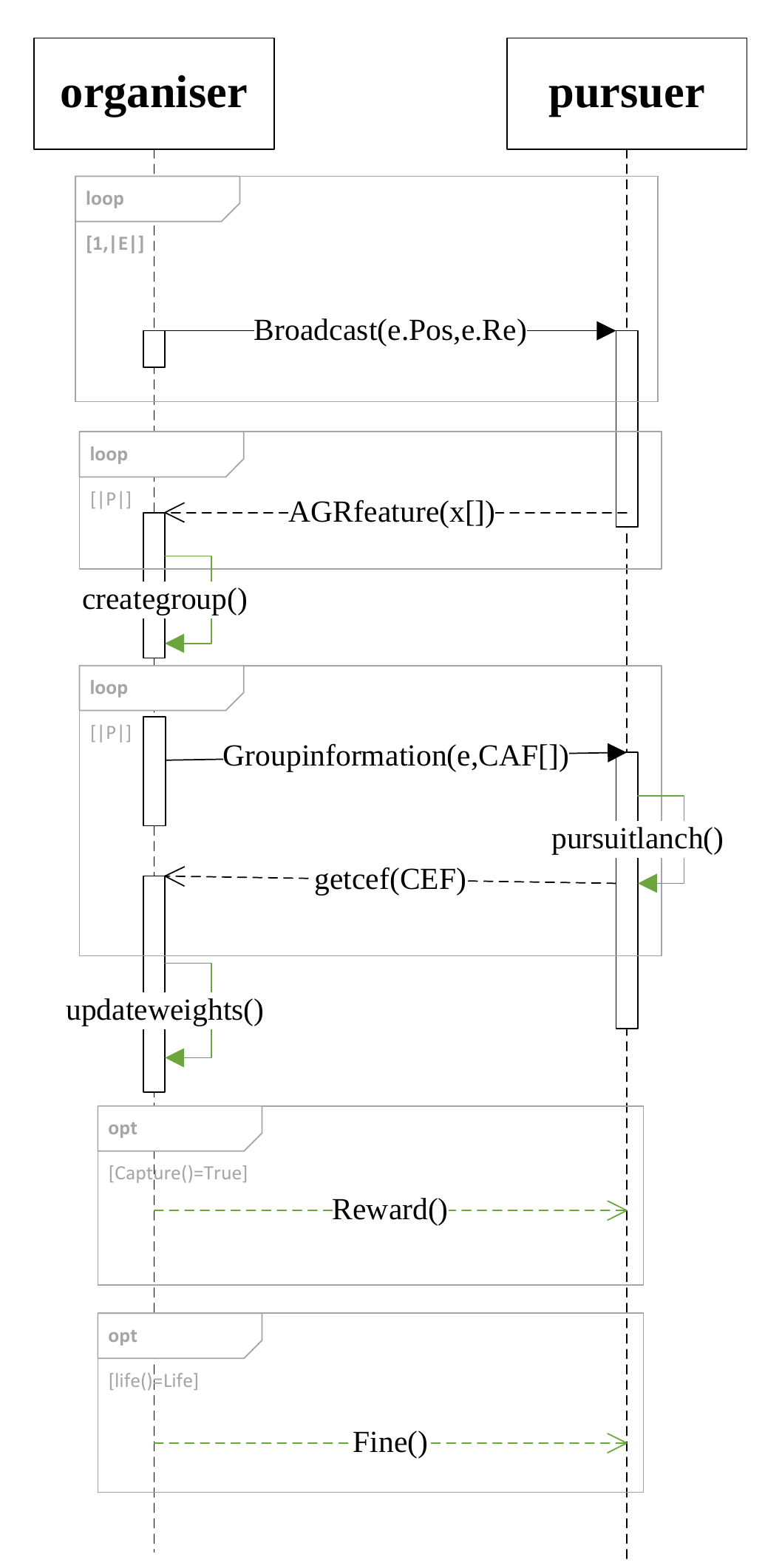}
  \caption{sequences diagram of communications}
  \label{shixu}
\end{figure}

\begin{algorithm}
  \small
  \scriptsize
  \caption{complete algorithm}
  \begin{algorithmic}[1]
    \State initialize the neuron network
    \Repeat
    \ForEach {$e \in E$}
    \State Broadcast (e.Pos, e.Re);
    \State Waiting-Response
    \EndFor
    \ForEach {$p \in P$}
    \State SendResponse($\overrightarrow{x_p}$)
    \EndFor
    \State GetResponse(X)
    \State train the SOM layer
    \State create-group()
    \State SendMessage(e,$\overrightarrow{GAF}$)
    \State Waiting-Response()
    \ForEach {$p \in P$}
    \State SendResponse(CEF)
    \If {$p.e\neq e $}
    \State p.life = 0
    \EndIf
    \State $p.e \gets e$
    \State Launch of the chase
    \State p.life = p.life + 1
    \If {$e.captured = true$}
    \State $p.C_t \gets p.C_t + 1$
    \State $p.C_s \gets p.C_s + 1$
    \EndIf

    \If {$p.life = life$}
    \State $p.C_t \gets p.C_t + 1$
    \State $p.C_s \gets p.C_b + 1$
    \EndIf
    \EndFor
    \State GetRespense($\overrightarrow{CEF}$))
    \State Train the feature extraction part
    \Until $\forall e \in E$ $e.captured=true$
  \end{algorithmic}
\end{algorithm}
The coalitions are short-lived and goal-oriented\cite{Souidi}.If a coalition
lasts longer than the default requirement $life$, it will be  considered as a failure.
The algorithm’s shown in Algorithm 3 are explained in the following manner:The
organizer initializes the neuron network(01).then the pursuers start to catch
evaders under the coordination of organizers until all evaders are
captured(02-33).The organizer broadcast position and value of each evader found
in environment and wait the response from pursuers(03-06).The pursuers response
with their own feature vector $\overrightarrow{x_p}$(07-09).The organizer train
the SOM layer with the set of feature vectors $X$ and send target and
$\overrightarrow{GAF}$ to each pursuer after creating group $p$(10-14).The
pursuer p response their $CEF(p)$ and start get close to its target(15-30).The
organizer train its feature extraction part with $\overrightarrow{CEF}$(31-32).The sequences diagram describing orginaser and pursuer communications is shown in Figure\ref{shixu}.

\section{Simulation Experiment}
\par Some experiments was done to verify the effectiveness of our overall
algorithm, that of feature extraction part and that of group generation part
separately.The pursuit environment in these simulations is a grid environment
with 100 *100 grids, with 16 pursuers and 2 evader with d = 4,1 evader with d =
3,1 evader with d = 2,The evaders disappeared after being captured and the
pursuers who captured them disappeared as well.we choose the algorithm based on
principles of MDP which was proposed in literature \cite{Souidi} as motion
strategy of pursuers.
\par \textbf{Experiment 1.}This experiment Verify the feasibility of coalition
formation of our algorithm based on AGRMF.The Figure \ref{Cct} shows the result of three
different ways which are explained below after 100 experiments for each case:
\begin{figure}
  \centering
  \includegraphics[width=0.6\linewidth]{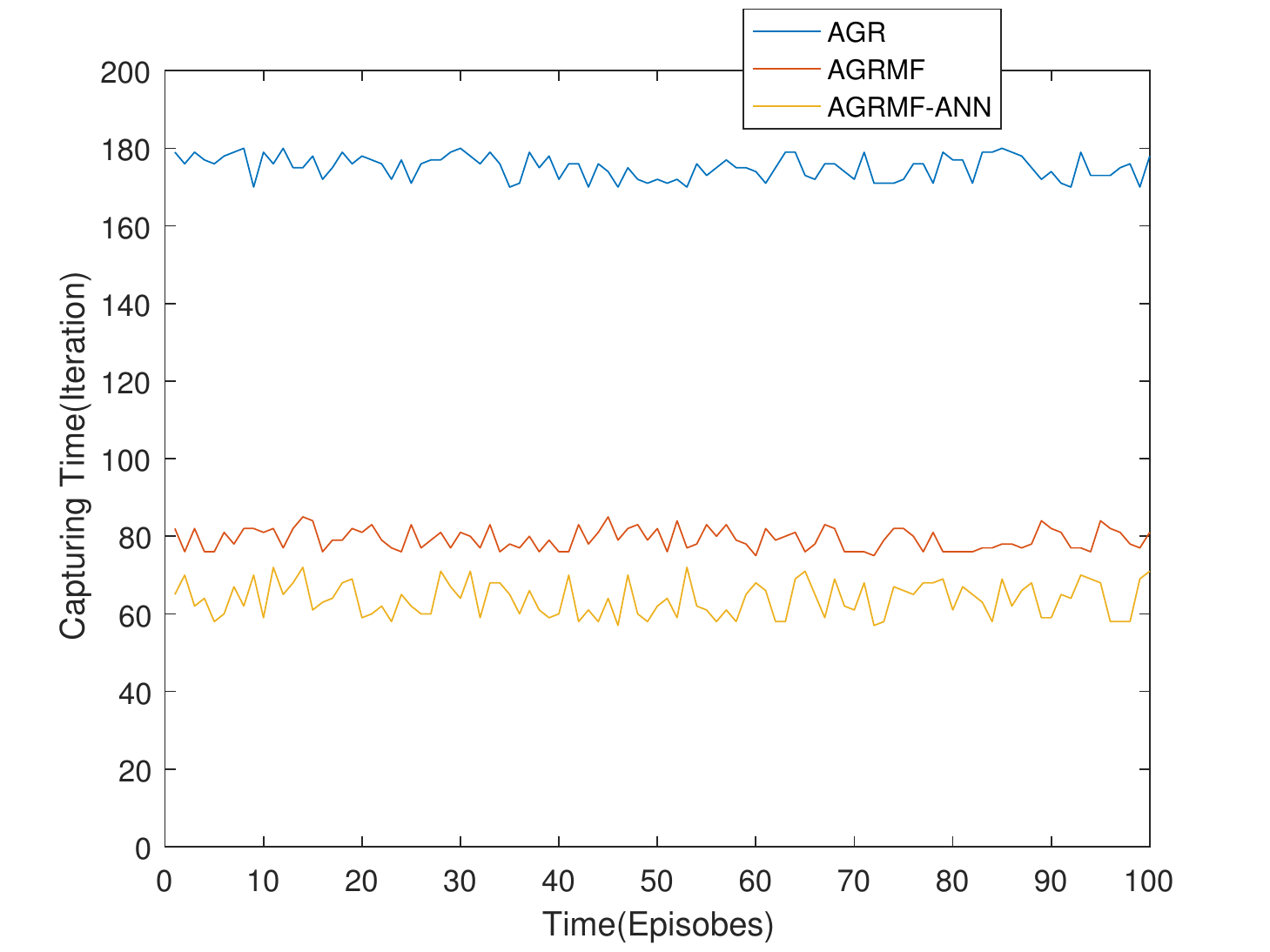}
  \caption{Comparison of capture time}
  \label{Cct}
\end{figure}

\begin{itemize}
\item Case AGR : formation of the groups without mechanism integration (without
membership function) based on AGR model refered to in literature \cite{Souidi}
\item Case AGRMF : formation and reformation of the groups with the application of the
  coalition algorithm which was proposed in literature \cite{Souidi} before any new chase iteration
\item Case AGRMF-ANN : coalition formation with the application of our coalition
  algorithm based on case AGRMF.
\end{itemize}
\par In case AGR, the capture of all evaders has consumed an average of 175.15 chase
iterations. For case AGRMF, capture of
evaders is performed after an average of 79.32 chase iterations. for case AGRMF-ANN,
capture of the evaders is achieved after an
average of 63.81 chase iterations.on the other hand,the error $E$ of feature
extraction of AGRMF-ANN's when it converges or stops its learning process for
each iteration is shown as figure \ref{lossfunction},when the evaders are caught
or when pursuers changes their targets,the error drops quickly,this part proves
that our algorithm for reorganization can promote pursuers pursuit the targets which can cause the
error reduce from a new state of AGRMF-ANN.
\begin{figure}
  \centering
  \includegraphics[width=0.6\linewidth]{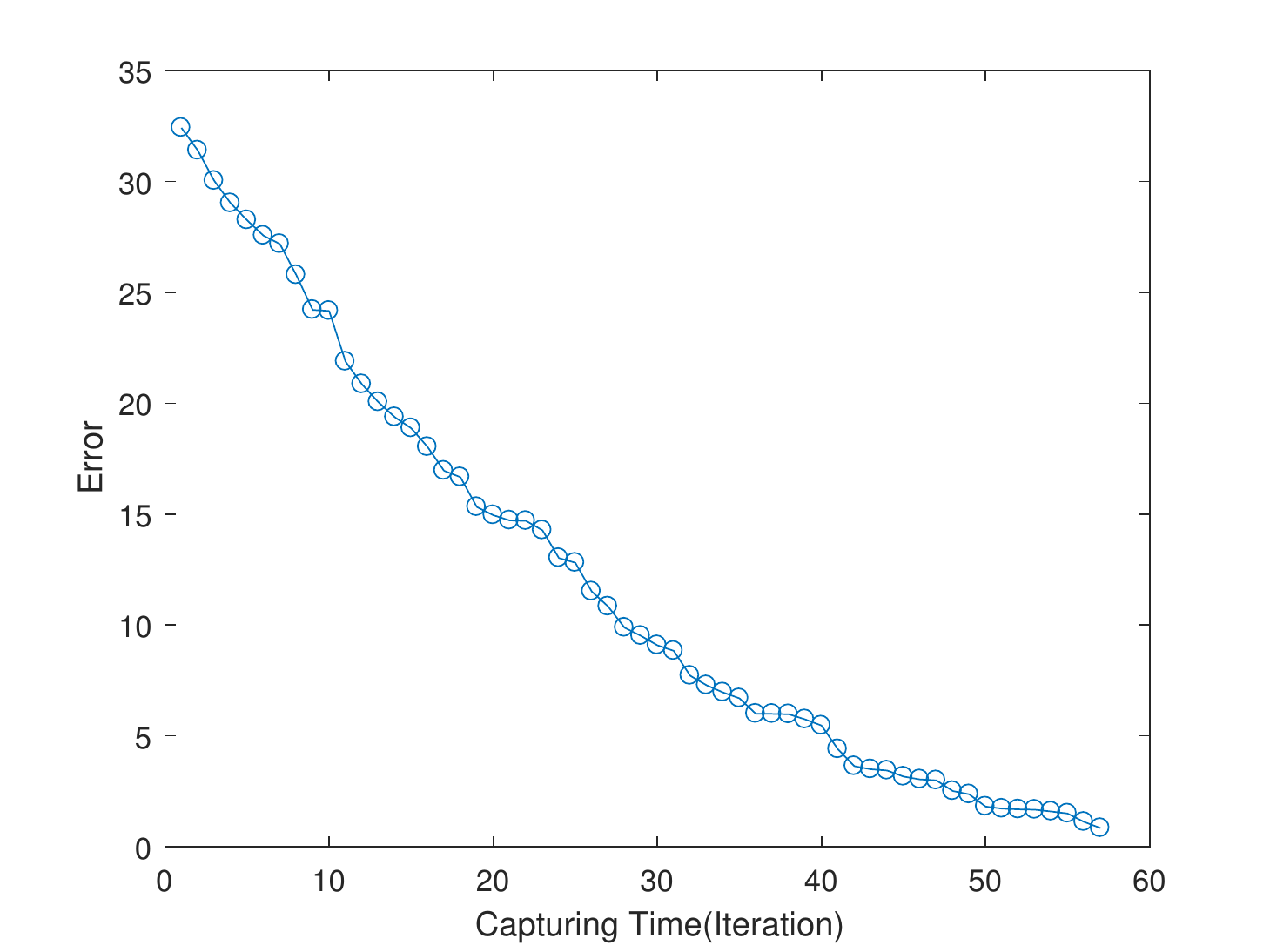}
  \caption{development of the error}
  \label{lossfunction}
\end{figure}
\par \textbf{Experiment 2.}This experiment mainly test the function of feature
extraction part of AGRMF-ANN according to comparison between two cases:
\begin{itemize}
\item case 1 : AGRMF-ANN without feature extraction part which means main ability
indicators should never change and group generation part creates coalitions
according to AGRMF directly.
\item case 2 : AGRMF-ANN.
\end{itemize}
The average chase iterations for case 1 is 94.650.
\begin{figure}
  \centering
  \includegraphics[width=0.6\linewidth]{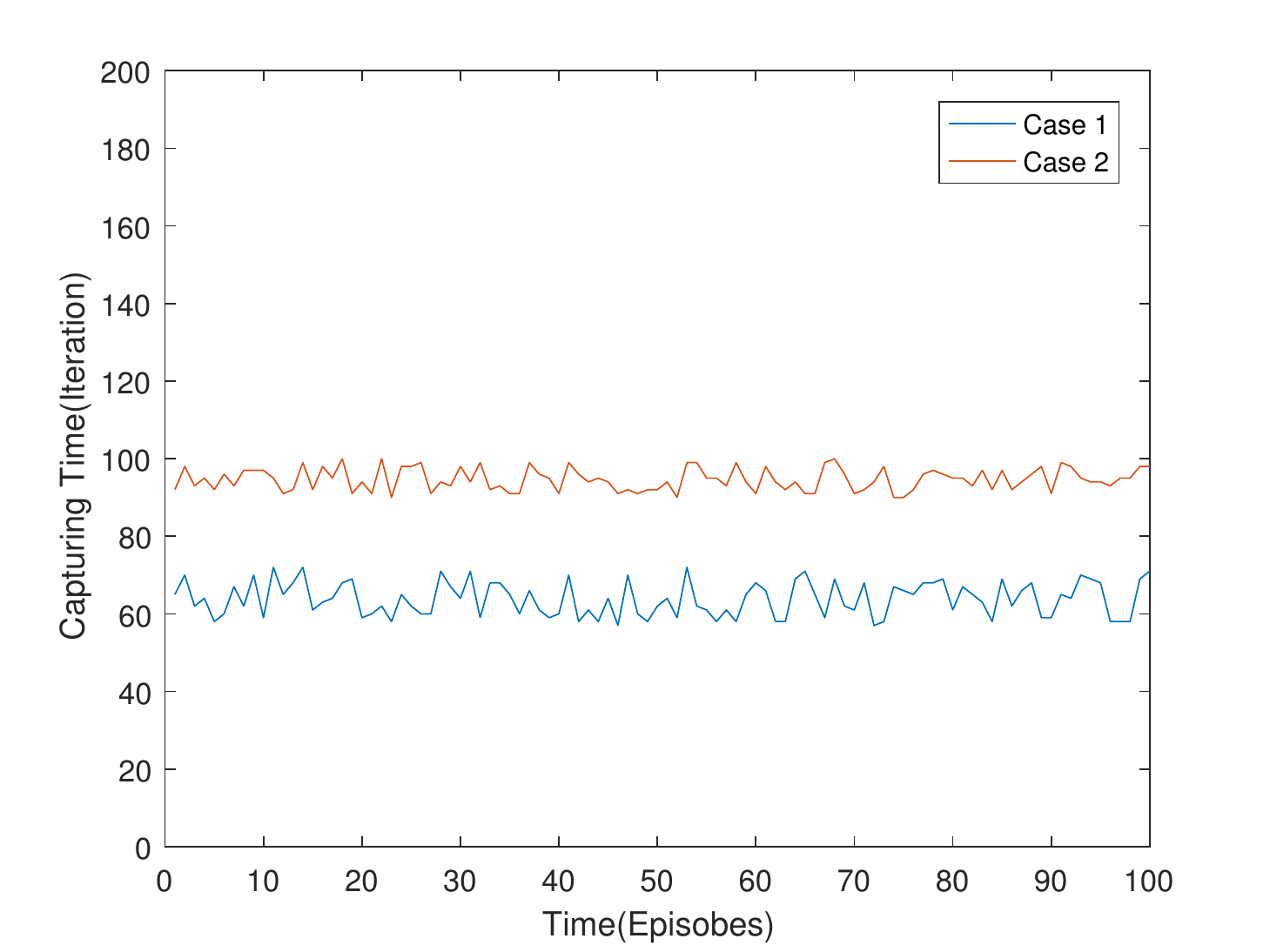}
  \caption{capture time of feature extraction inspired by different algorithms}
  \label{featureextraction}
\end{figure}
\par
\begin{figure}
  \centering
  \includegraphics[width=0.6\linewidth]{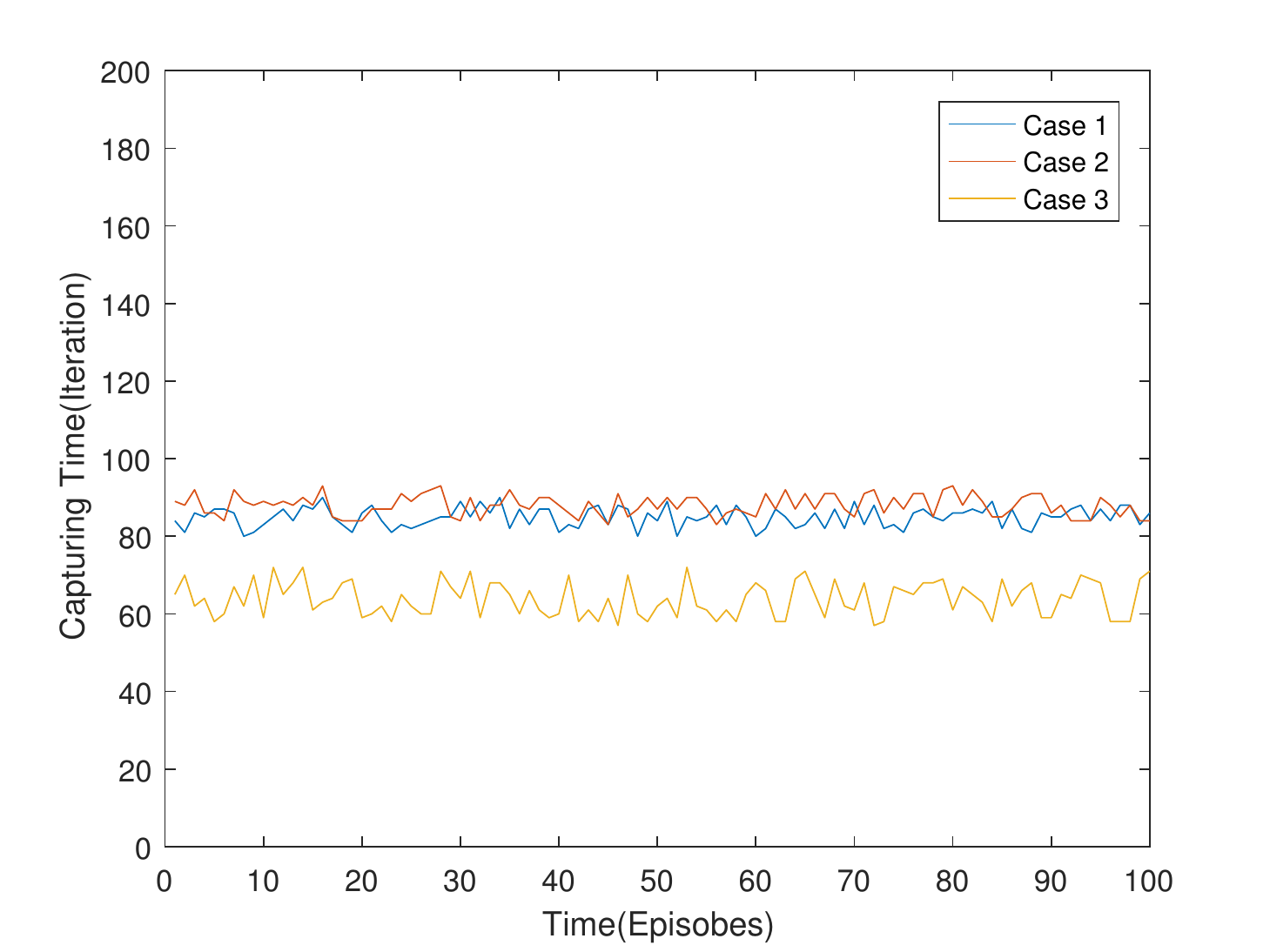}
  \caption{capture time of group generation inspired by different approaches}
  \label{cluster}
\end{figure}

 \textbf{Experiment 3.}We pick up KMEANS and DBSCAN which represent the cluster
algorithms based on partition and density respectively as contrasting objects to
explain the advantage of group generation part inspired by our algorithm.These
approaches all have their own advantage for data set with different characteristic.\cite{Viet} However,for this kind of
problem, the algorithm inspired by SOM perform best as shown in fig. \ref{cluster}.the
result shows the density and the number of centers are changing
dynamically during the process of chase, the competitive neuron network has a better ability to adapt to this change.

\par
\begin{figure}
  \centering
  \includegraphics[width=0.6\linewidth]{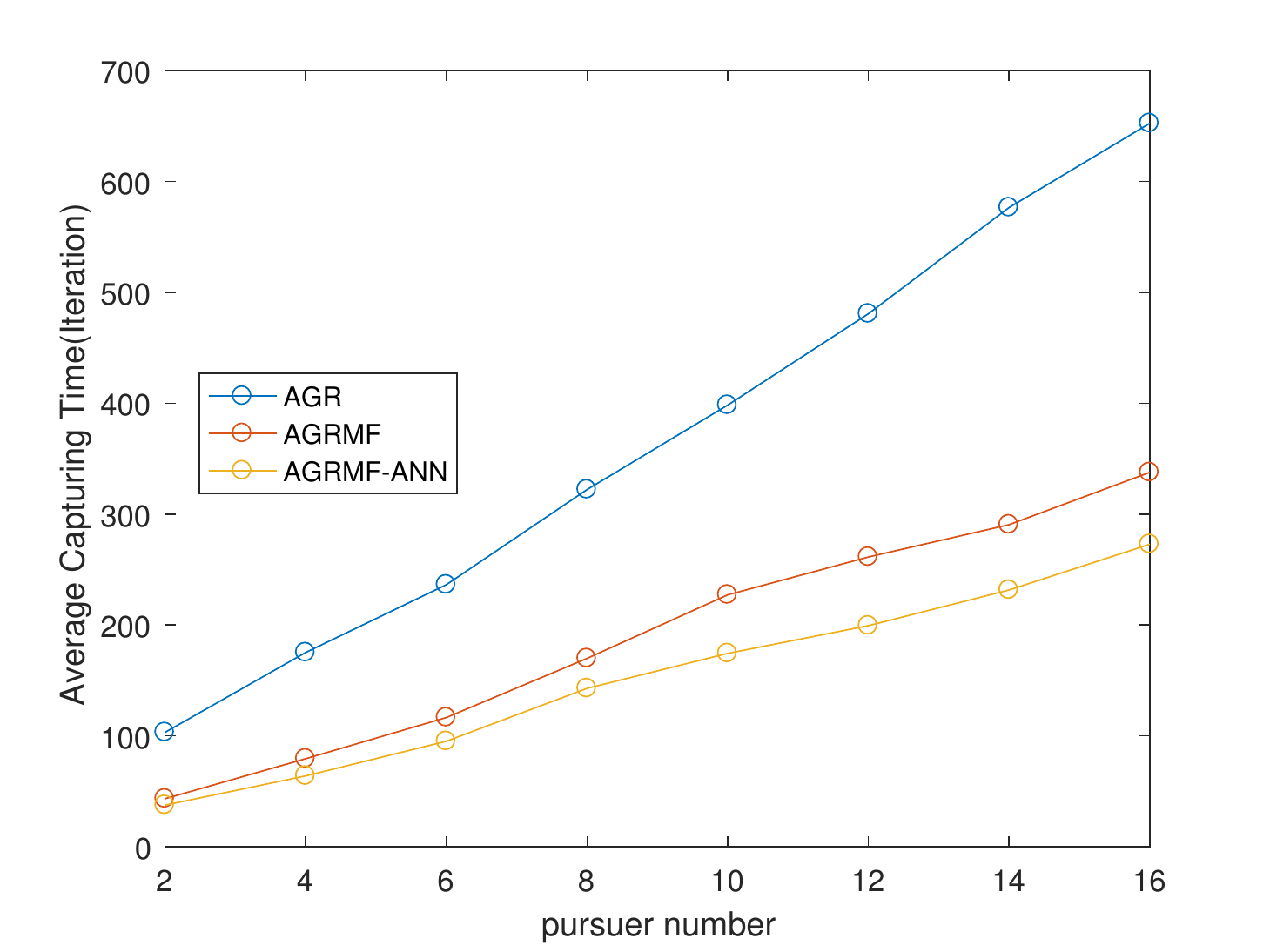}
  \caption{development of capture time}
  \label{pursuernum}
\end{figure}

\begin{figure}
  \centering
  \includegraphics[width=0.6\linewidth]{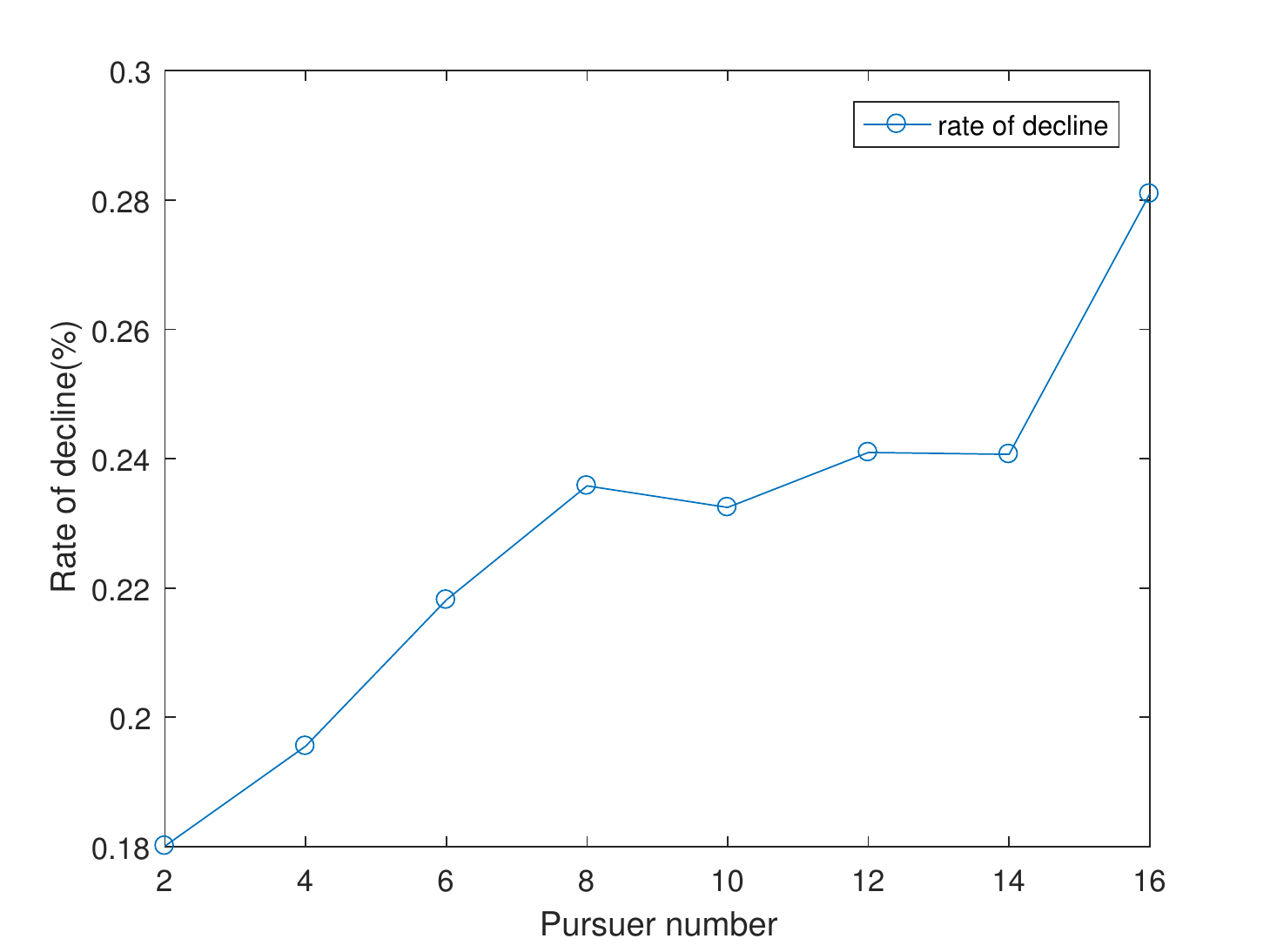}
  \caption{rate of decline compared with AGRMF}
  \label{declinerate}
\end{figure}

 \textbf{Experiment 4.}In order to validate our algorithm's ability to adapt
to larger environments, we designed experiment 4.the number of pursuers and
evaders scaled up as well as the size of environment on the basis of the
previous environment. the
capture time of the three case is shown as figure \ref{pursuernum}
 Growth of rate of decline for capture time of case AGRMF-ANN compared with case
 AGRMF can be represented by a linear shown in figure \ref{declinerate}.Because
 AGRMF - ANN can learn the attributes of some agents in the previous environment, it has advantages in the face of larger environments.

\section{Conclusion}
In this paper coalition formation for multi-agent pursuit based a novel neuron
network called AGRMF-ANN consisting of feature extraction part and group
generation part for to learn main ability indicators of AGRMF model is
discussed.Besides,learning algorithm of AGRMF based on group attractiveness
function(GAF) is also proposed.The simulation results shows effect on the
capture time comparing with AGRMF and the two part effectiveness.Next,we will
work to solve excessive load of orginaser because the entire network is deployed
on it and try to design an approach based on AGRMF-ANN can be applied for the environment of continuous space.
\section*{Funding}
This paper is supported by National Natural Science Foundation of China [grant number 61375081]; a
special fund project of Harbin science and technology innovation talents research [grant number
RC2013XK010002].

\end{document}